\documentclass{article}

\PassOptionsToPackage{numbers, compress}{natbib}

\usepackage[preprint]{neurips_2026}

\usepackage[utf8]{inputenc} 
\usepackage[T1]{fontenc}    
\usepackage{url}            
\usepackage{booktabs}       
\usepackage{amsfonts,amsmath,bbm}       
\usepackage{nicefrac}       
\usepackage{microtype}      
\usepackage{svg}
\usepackage{tikz}
\usepackage{multirow}

\usepackage{pgfplots}
\usepackage{siunitx}
\pgfplotsset{compat=1.18}
\usepackage{caption,subcaption}
\usepackage{tabularx}
\usepackage{graphicx}

\usepackage{pifont}
\usepackage{xcolor}
\usepackage{colortbl}
\usepackage{epstopdf}
\pgfplotsset{compat=1.18}

\usepackage{listings}
\usepackage{algorithm, algpseudocode}

\definecolor{nipsred}{rgb}{0.65,0.17,0.17}
\definecolor{lightgray}{rgb}{0.9,0.9,0.9}
\usepackage[pagebackref,breaklinks,colorlinks,allcolors=nipsred,urlcolor=black]{hyperref}
\usepackage{cleveref}

\renewcommand{\paragraph}[1]{\vspace{0em}\noindent\textbf{#1}}

\usepackage{wrapfig}

\title{Probing Diffusion Denoising Dynamics for Contrastive Representation Learning}

\author{
Yasong Dai\textsuperscript{\rm 1,2},
Zeeshan Hayder\textsuperscript{\rm 1,2},
David Ahmedt-Aristizabal\textsuperscript{\rm 2},
Hongdong Li\textsuperscript{\rm 1,3} \\
\textsuperscript{\rm 1}Australian National University, \textsuperscript{\rm 2}CSIRO Data61, \textsuperscript{\rm 3}Amazon \\
{\tt\small \{yasong.dai, zeeshan.hayder, hongdong.li\}@anu.edu.au}\\
{\tt\small \{david.ahmedtaristizabal\}@data61.csiro.au}
}

\begin{document}

\maketitle

\begin{abstract}
Text-to-image diffusion models exhibit unprecedented generative capability and contain rich intermediate representations that can be useful for discriminative vision tasks. Motivated by this observation, we study a focused question: how can the denoising dynamics of a pretrained diffusion model be adapted to support discriminative representation learning while preserving its generative behavior under parameter-efficient updates?
We present D$^3$CL as an investigation of this question. Our key observation is that noisy latents at different diffusion timesteps can be interpreted as stochastic views of the same underlying image, enabling a contrastive objective to be coupled with the standard denoising reconstruction loss. This formulation provides a simple way to probe the interaction between generative denoising and discriminative representation learning without training from scratch. To keep the adaptation lightweight, we apply LoRA updates to a pretrained Stable Diffusion backbone while freezing the original model parameters. D$^3$CL provides strong empirical evidence that reconstruction and noise-level contrastive objectives can be complementary: on ImageNet-1K, it obtains 80.1\% linear-probing accuracy and an FID of 5.56 for $256 \times 256$ unconditional generation. Additional ablations on the design space suggest that the usefulness of diffusion features depends on where and how denoising states are sampled.
These results establish D$^3$CL as a parameter-efficient adaptation framework for pretrained diffusion models, showing that noise-level contrastive learning can structure denoising representations for discriminative tasks while maintaining generative performance.
\end{abstract}
\section{Introduction}
\label{sec:intro}

Self-supervised representation learning has demonstrated remarkable results in deriving rich, transferable features without additional supervision signals. Contrastive approaches~\citep{chen2020simclr,caron2021dino} and generative methods~\citep{he2022mae,xie2022simmim} have been developed along separate paths to learn robust visual representations. However, recent research~\citep{mukhopadhyay2023diffeed, park2023vitlearn} suggests that both contrastive and generative paradigms have shared underlying principles in capturing semantic information from unlabeled data.

Following this idea, several methods~\citep{li2023mage,hudson2024soda, zhu2024digit} have aimed to unify self-supervised learning for both generative and discriminative tasks. However, these methods still encounter notable limitations, particularly in balancing the trade-off between feature robustness for recognition and high-quality generation~\citep{he2022mae}.
Another challenge arises largely from the extensive computational demands. A state-of-the-art model~\citep{li2023mage}, for example, relies on a heavily parameterized ViT-L/16 backbone with over 400M trainable parameters, requiring 1600 epochs of training. This high resource demand limits the practicality of such models in real-world applications. This raises a critical research question in self-supervised representation learning: {\bf Can we develop a unified framework that effectively balances feature robustness and generation quality while being computationally efficient?}

Remarkable advancements in generative models present a promising direction for the question. Diffusion models, in particular, have emerged as a powerful framework for high-fidelity image generation~\citep{ho2020ddpm} and meaningful representation learning~\citep{preechakul2022diffae,mittal2023drl}, suggesting a unique opportunity to unify generative and discriminative tasks under a single framework.
Modern Stable Diffusion models are pre-trained on large scale datasets~\citep{schuhmann2022laion} and open-source, making fine-tuning and fast adaptation on them efficient without the need for training from scratch.

\definecolor{dinocol}{RGB}{170,70,175}
\definecolor{diffeedcol}{RGB}{255,190,205}
\definecolor{ibotcol}{RGB}{255,70,55}
\definecolor{simclrcol}{RGB}{210,45,255}
\definecolor{maecol}{RGB}{55,55,255}
\definecolor{admcol}{RGB}{45,235,235}
\definecolor{bigbigancol}{RGB}{245,240,40}
\definecolor{maskgitcol}{RGB}{190,80,80}
\definecolor{givtcol}{RGB}{65,135,65}
\definecolor{icgancol}{RGB}{65,255,65}
\definecolor{magecol}{RGB}{255,180,55}
\definecolor{contracol}{RGB}{165,165,165}

\begin{figure}
\centering
\begin{tikzpicture}
\begin{axis}[
    width=10cm,
    height=7.2cm,
    xmin=0, xmax=38,
    ymin=58, ymax=82,
    x dir=reverse,
    x axis line style={-},
    axis lines=left,
    clip=false,
    xlabel={FID (Unconditional Generation) $\downarrow$},
    ylabel={Linear Probing Accuracy (\%) $\uparrow$},
    xlabel style={font=\normalsize, yshift=-6pt},
    ylabel style={font=\normalsize, yshift=6pt},
    tick label style={font=\normalsize},
    xtick={35,30,25,20,15,10,5,0},
    xticklabels={N/A,30,25,20,15,10,5,0},
    ytick={60,65,70,75,80},
    yticklabels={N/A,65,70,75,80},
    major grid style={dashed, gray!55},
    grid=major,
    tick style={black},
    axis line style={black, thick},
]


\addplot[
    only marks,
    mark=*,
    mark size=6.5pt,
    mark options={fill=dinocol, draw=none, fill opacity=0.75}
] coordinates {(35,78.0)};
\node[anchor=west, font=\normalsize] at (axis cs:33.85,78.0) {DINO};

\addplot[
    only marks,
    mark=*,
    mark size=3.5pt,
    mark options={fill=diffeedcol, draw=none, fill opacity=0.55}
] coordinates {(35,76.6)};
\node[anchor=west, font=\normalsize] at (axis cs:33.85,76.6) {DiffFeed};

\addplot[
    only marks,
    mark=*,
    mark size=5.0pt,
    mark options={fill=ibotcol, draw=none, fill opacity=0.75}
] coordinates {(35,75.0)};
\node[anchor=west, font=\normalsize] at (axis cs:33.85,75.0) {iBOT};

\addplot[
    only marks,
    mark=*,
    mark size=13pt,
    mark options={fill=maecol, draw=none, fill opacity=0.70}
] coordinates {(35,72.5)};
\node[anchor=north, font=\normalsize] at (axis cs:35,71.0) {MAE};

\addplot[
    only marks,
    mark=*,
    mark size=6.5pt,
    mark options={fill=simclrcol, draw=none, fill opacity=0.75}
] coordinates {(35,73.4)};
\node[anchor=west, font=\normalsize] at (axis cs:33.85,73.6) {SimCLR};


\addplot[
    only marks,
    mark=*,
    mark size=13pt,
    mark options={fill=admcol, draw=none, fill opacity=0.70}
] coordinates {(25.8,60.0)};
\node[anchor=east, font=\normalsize] at (axis cs:27.7,60.0) {ADM};

\addplot[
    only marks,
    mark=*,
    mark size=3.6pt,
    mark options={fill=bigbigancol, draw=none, fill opacity=0.85}
] coordinates {(22.8,61.3)};
\node[anchor=south, font=\normalsize] at (axis cs:22.8,62.0) {BigBiGAN};

\addplot[
    only marks,
    mark=*,
    mark size=10.5pt,
    mark options={fill=maskgitcol, draw=none, fill opacity=0.65}
] coordinates {(20.3,60.0)};
\node[anchor=west, font=\normalsize] at (axis cs:18.7,60.0) {MaskGIT};

\addplot[
    only marks,
    mark=*,
    mark size=5.5pt,
    mark options={fill=icgancol, draw=none, fill opacity=0.75}
] coordinates {(11.0,60.0)};
\node[anchor=west, font=\normalsize] at (axis cs:9.9,60.0) {ICGAN};

\addplot[
    only marks,
    mark=*,
    mark size=12pt,
    mark options={fill=givtcol, draw=none, fill opacity=0.75}
] coordinates {(11.0,65.1)};
\node[anchor=west, font=\normalsize] at (axis cs:9.3,65.1) {GIVT};


\addplot[
    only marks,
    mark=*,
    mark size=15.5pt,
    mark options={fill=magecol, draw=none, fill opacity=0.75}
] coordinates {(6.2,78.9)};
\node[anchor=west, font=\normalsize] at (axis cs:4.9,76.9) {MAGE};

\addplot[
    only marks,
    mark=*,
    mark size=6pt,
    mark options={fill=contracol, draw=none, fill opacity=0.85}
] coordinates {(5.0,80.1)};
\node[anchor=west, font=\normalsize] at (axis cs:3.9,80.1) {D$^3$CL(\emph{ours})};

\end{axis}
\end{tikzpicture}
    \caption{{\bf D$^3$CL balances accuracy and efficiency.} We report linear probing and unconditional image generation performance of different methods on ImageNet-1K. The area of a circle corresponds to the number of trainable parameters. Our method outperforms baseline models in both discriminative (classification) and generative (unconditional image generation) tasks, even surpassing those trained for only one of these tasks. In the meantime, our method maintains a small number of trainable parameters to reduce training resource overhead.}
    \label{fig:intro_overview}
\end{figure}
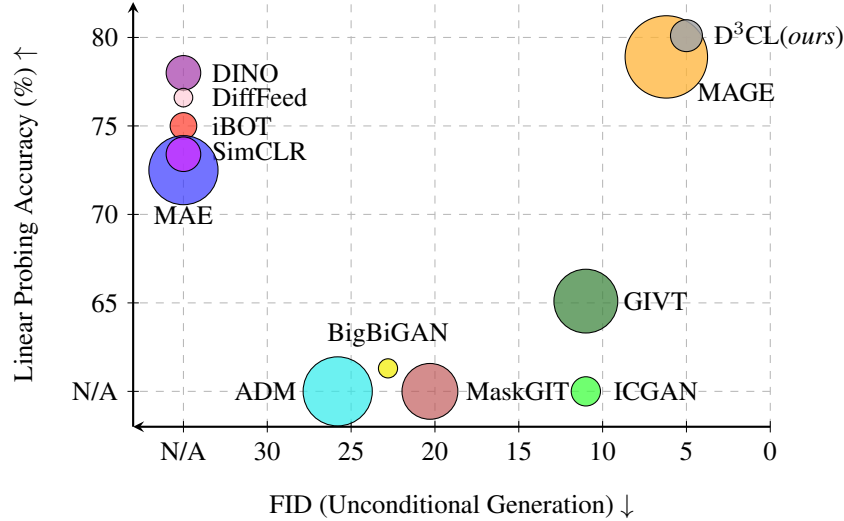

In this work, we propose \textsc{D$^3$CL}, a novel framework that integrates representation learning and generative modeling within a single diffusion process. Our key technical novelty is the incorporation of contrastive learning into diffusion models: In the reverse diffusion process, where images are progressively denoised through sequential steps, contrastive loss can be naturally applied by treating images at different noise levels as distinct ``views'' of the same underlying data. Inspired by SimCLR~\citep{chen2020simclr}, we incorporate a contrastive objective that operates across varying noise levels, leveraging both the efficiency and discriminative benefits of contrastive learning. This enables D$^3$CL to learn robust features for discriminative tasks while preserving its ability to generate high-fidelity images.

To address the high computational demands inherent to large-scale diffusion models, we integrate LoRA~\citep{hu2021lora} as an efficient adaptation mechanism. 
Specifically, we apply LoRA to the cross-attention matrices in Stable Diffusion during training, enabling efficient UNet weight updates that align with the image condition latent with minimal computational cost. By reducing resource requirements, D$^3$CL allows for the simultaneous application of representation learning and generative modeling within a unified framework, reducing adaptation cost while preserving generative quality.

Our framework demonstrates competitive classification accuracy and high-quality image generation on ImageNet-1K~\citep{russakovsky2015imagenet}, outperforming certain task-specific contrastive methods. Through comprehensive empirical evaluation, we highlight the effectiveness of unifying contrastive and generative learning, showing that these approaches can coexist within a single framework to yield strong results across both classification and image synthesis tasks. In summary, our main contributions are as follows: \textbf{(1) A novel framework} that bridges representation learning and generative modeling by learning contrastive features obtained from generative denoising steps in diffusion processes, boosting both image generation and classification performance.
\textbf{(2) Comprehensive empirical evaluation} demonstrating D$^3$CL's strong image generation capabilities alongside high classification accuracy. Additionally, transfer learning experiments on CIFAR-100 confirm the generalization ability of our method.
\section{Related Work}
\label{sec:literature}

\paragraph{Self-supervised learning in recognition tasks.}
Self-supervised learning has transformed computer vision by enabling models to learn from unlabeled data using its inherent structure to create supervision signals. Early advances in the area were driven by contrastive methods, where models learn meaningful representations by contrasting positive and negative sample pairs. Pioneering methods like SimCLR~\citep{chen2020simclr} and MoCo~\citep{he2020moco} maximize similarity between different views of the same image, contrasting these with other images. Later, non-contrastive approaches such as DINO~\citep{caron2021dino} introduced a teacher-student self-distillation approach, where the student model learns to match representations from a teacher network. Collectively, these methods have shown that contrastive and distillation-based self-supervision can learn high-quality representations without labeled data. 

However, most early self-supervised learning methods~\cite{caron2021dino,he2022mae} require extensive pretraining to reach competitive performance, often training from scratch over hundreds of epochs. 
Furthermore, while generative models like MAE~\citep{he2022mae} demonstrate promising reconstruction abilities, they often struggle to balance image fidelity with robust feature learning, especially in high-fidelity generative tasks. Consequently, there is a pressing need for methods that unify robust feature extraction with high-quality generation within a more resource-efficient, self-supervised framework.

\paragraph{Diffusion model for discriminative tasks.}
Diffusion models~\citep{sohl2015deep,ho2020ddpm} are a class of generative models that progressively convert random noise into high-fidelity image samples. In addition to recent works~\citep{saharia2022imagen,ramesh2022dalle2,rombach2022ldm} that achieved remarkable results in high-quality and diverse image synthesis, their potential for representation learning has gained attention due to their ability to capture rich, hierarchical features. 
DiffAE~\citep{preechakul2022diffae} uses an auto-encoding process within the diffusion framework, effectively reconstructing input data from noise to capture meaningful latent features. DiffMAE~\citep{wei2023diffmae} combines diffusion with masked autoencoders, enhancing feature extraction and generalization by reconstructing partially corrupted inputs. Diffusion Classifier~\citep{li2023diffusionclassifier} further extends diffusion models to classification tasks.

Adapting diffusion models for self-supervised learning still presents challenges. These models are inherently large~\citep{rombach2022ldm,karras2022elucidating}, making full fine-tuning computationally expensive. Additionally, current approaches to feature extraction, such as DifFeed~\citep{mukhopadhyay2023diffeed} and DDAE~\citep{xiang2023ddae}, often depend on frozen, pretrained diffusion models. This limits flexibility when extending to other discriminative tasks, as frozen models may not adapt effectively across different contexts~\citep{mukhopadhyay2023diffeed}. 

\paragraph{Unified self-supervised learning for discriminative and generative tasks.}
Recent advancements in unified self-supervised learning frameworks aim to support both discriminative and generative tasks within a single model, reflecting a shift towards versatile, efficient learning paradigms. MAGE~\citep{li2023mage} introduces a self-supervised approach that learns joint representations for both tasks via a novel masking strategy and a contrastive loss. However, MAGE requires an extensive pretraining phase to achieve robust representations, making it resource-intensive. For diffusion models, SODA~\citep{hudson2024soda} employs a compact bottleneck to the representation from its DDPM~\citep{ho2020ddpm} conditional encoder, training separate encoder and generator modules for unified task execution. Despite these advances, existing frameworks often depend on heavy pretraining and substantial computational resources, which limit their adaptability. This underscores the need for a resource-efficient unified framework capable of high performance in both discriminative and generative tasks with minimal computational overhead.
\section{Method}
\label{sec:method}

\subsection{Preliminaries for diffusion models}
Diffusion models have emerged as a powerful class of generative models, known for their ability to generate high-quality images by modeling the data generation process as a reverse diffusion process.

\paragraph{Forward process.} 
A diffusion model operates through a sequence of gradual, noise-adding transformations that convert data from a complex distribution into a simpler distribution (e.g., a Gaussian distribution) over a predefined number of steps. This process is inspired by non-equilibrium thermodynamics~\citep{sohl2015deep} and has been refined across the works of \citet{song2020score, ho2020ddpm,song2020ddim}. Formally, the diffusion forward process can be described by a discrete Markov chain in \Cref{eqn:forward}, where \(x_{t}\) represents noisy data at discrete time step \(t\), \(\beta_{t}\) is the variance schedule which controls the noise level at each step, progressively transforming the data into noise.
\begin{equation}\label{eqn:forward}
    q(x_{t} | x_{t-1}) = \mathcal{N}\left(\sqrt{1 - \beta_{t}}\; x_{t-1}, \beta_{t} I\right)
\end{equation}

\paragraph{Reverse process.} 
The reverse process, which is the core of a diffusion model's generative capability, aims to reconstruct the original data distribution $x_0 \sim p_\text{data}(x)$ from the noise. The DDPM reverse process is formalized as \Cref{eq:reverse}, where $\alpha_t := 1 - \beta_t, \bar{\alpha}_t:=\prod_{s=1}^t\alpha_s$, $\epsilon \sim \mathcal{N}(0,I)$, and \(\boldsymbol{\epsilon}_{\theta}(x_{t}, t)\) is a neural network that learns to predict the noise component with $x_{t}$ and $t$.
\begin{equation}\label{eq:reverse}
    x_{t-1} = \frac{1}{\sqrt{1 - \beta_{t}}} \left(x_{t} - \frac{\beta_{t}}{\sqrt{1 - \bar{\alpha}_{t}}} \boldsymbol{\epsilon}_{\theta}(x_{t}, t)\right) + \sqrt{\frac{1-\bar{\alpha}_{t-1}}{1-\bar{\alpha}_t} \beta_t} \cdot\epsilon
\end{equation}

\paragraph{Latent diffusion models (LDM).}
During training, LDMs first compress input images into a low-dimensional latent $z$ with a pre-trained visual encoder $\mathcal{E}$, then perform noise-adding and denoising in latent space, and decode reconstructed latent via a decoder $\mathcal{D}: \Tilde{x} = \mathcal{D}(\Tilde{z})$, where $z=\mathcal{E}(x)$. This compression procedure preserves semantic information of image data while being more efficient in terms of computational resources, as evidenced by \citet{rombach2022ldm}.

\subsection{Method Overview}

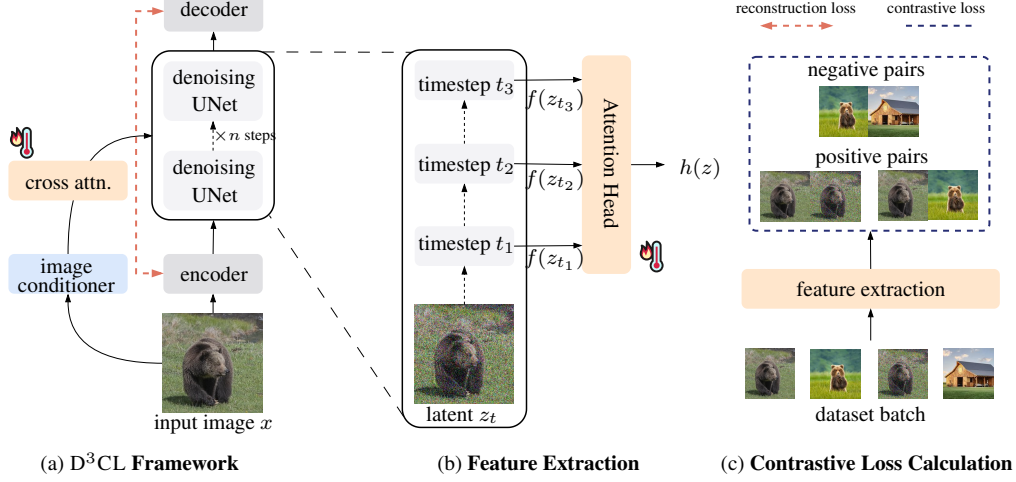
\begin{figure*}[t!]
    \centering
    \small
    \begin{tikzpicture}[every node/.style=
    {
    font=\footnotesize,
    align=left,
    anchor=south west
    }]

        \node[anchor=south west, inner sep=0] (image) at (0,0) {\includegraphics[width=\linewidth]{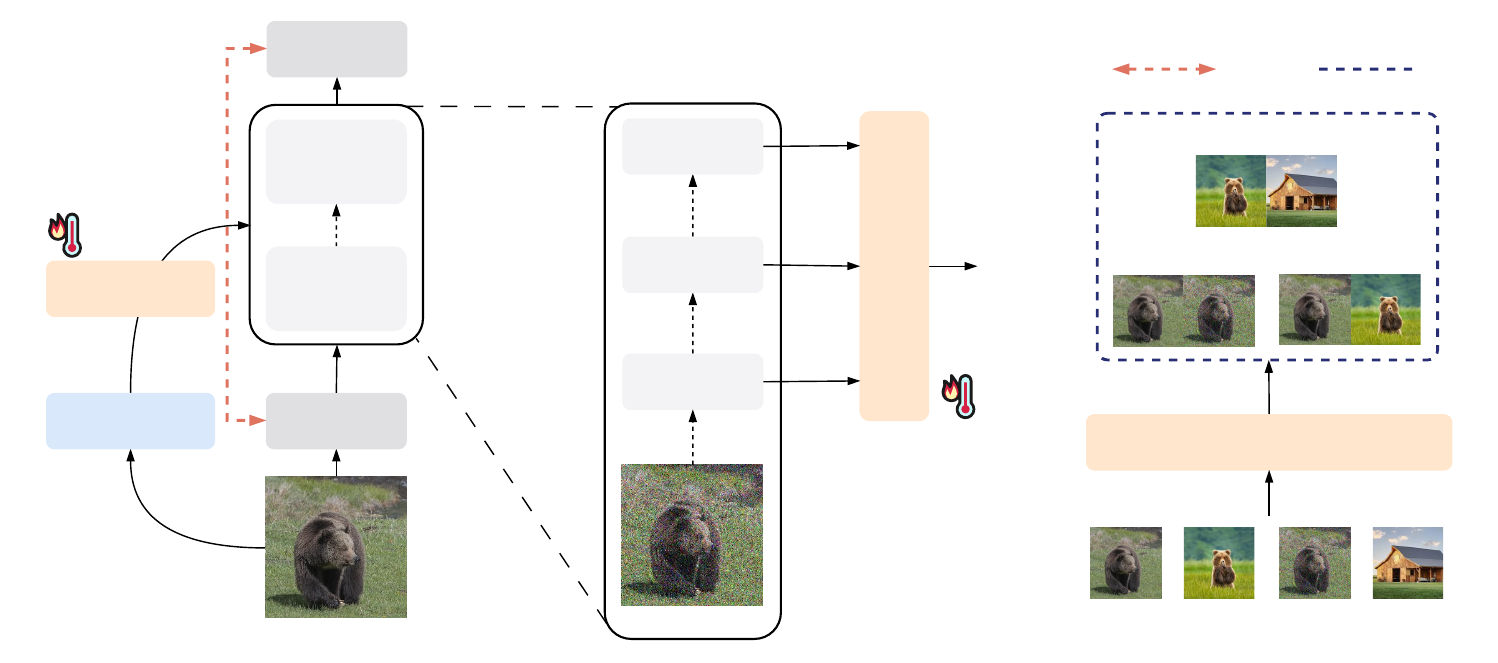}};
    
        \node[] at (2.25, 0.03) {input image \(x\)};
        \node[] at (2.60, 2.05) {encoder};
        \node[] at (2.60, 5.55) {decoder};
        \node[] at (0.75, 2.1) {image};
        \node[] at (0.45, 1.9) {conditioner};
        \node[] at (0.55, 3.27) {cross attn.};
        \node[] at (2.50, 4.60) {denoising};
        \node[] at (2.75, 4.30) {UNet};
        \node[] at (2.50, 3.40) {denoising};
        \node[] at (2.75, 3.10) {UNet};
        \node[] at (3.03, 3.87) {\tiny \(\times n\) steps};
    
        \node[rotate=270] at (8.20, 4.70) {Attention Head};
        \node[] at (5.85, 0.17) {latent \(z_t\)}; 
        \node[] at (5.75, 2.40) {timestep \(t_1\)};
        \node[] at (7.15, 2.20) {\(f(z_{t_1})\)};
        \node[] at (5.75, 3.45) {timestep \(t_2\)};
        \node[] at (7.15, 3.25) {\(f(z_{t_2})\)};
        \node[] at (5.75, 4.50) {timestep \(t_3\)};
        \node[] at (7.15, 4.30) {\(f(z_{t_3})\)};
        \node[] at (9.20, 3.40) {\(h(z)\)};
    
        \node[] at (11.0, 0.20) {dataset batch};
        \node[] at (10.75, 1.85) {feature extraction};
        \node[] at (11.00, 3.55) {positive pairs};
        \node[] at (10.90, 4.70) {negative pairs};
    
        \node[] at (9.95, 5.60) {\tiny reconstruction loss};
        \node[] at (11.95, 5.60) {\tiny contrastive loss};
    
        \node[] at (0.75, -0.50) {(a) \textsc{D$^3$CL} {\bf Framework}};
        \node[] at (6.0, -0.50) {(b) \bf Feature Extraction};
        \node[] at (9.75, -0.50) {(c) \bf Contrastive Loss Calculation};
    
    \end{tikzpicture}
    
    \caption{{\bf Overview of the D$^3$CL training pipeline.} 
    An input image is encoded by a VAE encoder to produce a latent representation $z$, which is then perturbed with noise to form a noisy latent of level $t$. This noisy latent is processed by a denoising UNet with the conditional latent applied on cross-attention layers for $n$ steps. To enhance efficiency, LoRA is applied in the QKV (query, key, value) attention layers. This setup allows D$^3$CL to balance generative and discriminative tasks effectively while reducing training resource requirements. The output of UNet is then decoded by a VAE decoder, reconstructing the image from the latent representation.}
    \label{fig:method_detail}
\end{figure*}

\textsc{D$^3$CL} extends the capabilities of a pre-trained Stable Diffusion model beyond generative tasks through efficient fine-tuning and feature extraction for representation learning. As shown in \Cref{fig:method_detail}, the input image $x$ is first encoded into latent representations $z$ by a VAE latent encoder. An image conditioner also generates image-based conditional latent using $x$. Next, Gaussian noise of level $t$ is added to the image latent following \Cref{eqn:forward}, forming a noisy latent representation. The noisy latent, along with image condition embeddings, is then fed into the denoising UNet of the Stable Diffusion model, which reconstructs the latent representation before it is decoded back into pixel space.

To achieve efficient adaptation while preserving the pre-trained weights, we integrate Low-Rank Adaptation (LoRA) matrices~\citep{hu2021lora} within the cross-attention layers of the denoising UNet. This strategy facilitates flexible fine-tuning and enhances representation learning without incurring extensive computational costs. Detailed explanations of each component follow below.

\subsection{Training Objectives}

\paragraph{Generative training.}
For each input image $x$, we encode it into latent space: $z=\mathcal{E}(x)$. To retain the model's generative capabilities while adapting it to new tasks, we employ a reconstruction loss on the model's denoising output, following the LDM loss formulation. Our primary goal is learning to reconstruct noisy latent $z_t$, which is equivalent to predicting the noise added on image latent representations, as formulated in \Cref{eqn:ldm}:
\begin{equation}\label{eqn:ldm}
\mathcal{L}_\text{recon}=\mathbb{E}_{x\sim p_\text{data}, \epsilon \sim \mathcal{N}(0,I), t}\left[\left\|\epsilon-\epsilon_\theta\left(z_t, t\right)\right\|^2\right],
\end{equation}
where $x$ is the input image, $\epsilon$ represents noise sampled from a Gaussian distribution, and $z_t$ is the noisy latent, which can be obtained from the model by the forward process.

\paragraph{Contrastive feature extraction.} 
We leverage the rich representations within the diffusion model by extracting features from the bottleneck layer of the UNet architecture, where spatial resolution is minimized, and semantic information is densely encoded. Specifically, during a denoising step $t$, when an image latent $z$ is passed through the UNet $\epsilon_\theta\left(z_t, t\right)$, we use the activation $f(z_t)$ from the bottleneck layer as the feature. To further enhance the extracted features, we apply a cross-attention mechanism~\citep{vaswani2017attention} to the output of the bottleneck layer at different denoising timesteps $(t_1, t_2, t_3)$:
\begin{equation}
h(z) = \texttt{Attention}\left[W_Q f(z_{t_1}), W_K f(z_{t_2}), W_V f(z_{t_3})\right]
\end{equation}
where $W_Q$, $W_K$, and $W_V$ are learnable projection matrices for query, key, and value transformations, respectively. This strategy encodes features from different denoising steps, resulting in a representation enriched with consistent semantic information.

\paragraph{Contrastive loss design.}
Following the approach in \citet{li2023mage}, we apply a contrastive learning strategy to enhance the separability of diffusion features for improved performance on discriminative tasks. To construct positive/ negative pairs, we treat different noise levels as unique ``views'' of an image. Specifically, given a clean image $x$, we first encode it into latent $z$. Then, we generate two distinct ``views'' of $z$ by applying different noise levels in the forward diffusion process:
\begin{equation}
z_t \sim q(z_t | z), z_{t'} \sim q(z_{t'}|z)
\end{equation}
where $(t,t')$ are time steps sampled from a fixed schedule. We employ InfoNCE loss~\citep{oord2018infonce} to maximize the mutual information between features extracted from these noisy views:
\begin{equation}
\mathcal{L}_\text{contrast} = -\sum_{i=1}^{N}  \log \frac{2\cdot\exp(\text{sim}(h_i, h'_i)/\tau)}{\sum_{j=1}^{N} \mathbbm{1}_{j\neq i} \left(\exp(\text{sim}(h_i, h_j)/\tau)+\exp(\text{sim}(h_i, h'_j)/\tau)\right)}
\end{equation}
where $N$ is the batch size, $\text{sim}(\cdot,\cdot)$ represents cosine similarity, $\tau$ is a temperature parameter, $h_i$ represents feature extracted from the $i$th sample, and $h_j$ denotes negative sample features in the batch.

\subsection{Training Framework}

\paragraph{Overall objective.} 
Our training process combines both reconstruction and contrastive learning objectives to enhance high-quality image generation while simultaneously learning robust features for discriminative tasks. The overall training loss is formulated as \Cref{eq:loss}, where $\lambda$ is a reweighting parameter that balances the contributions of the reconstruction and contrastive objectives. We set $\lambda=0.1$ for the training process, chosen via grid search as shown in \Cref{tab:re_ablation}.
\begin{equation}\label{eq:loss}
\mathcal{L} = \mathcal{L}_\text{recon} + \lambda \times \mathcal{L}_\text{contrast}
\end{equation}

\paragraph{Noise schedule in diffusion process.}
Unlike standard sine or cosine noise schedules commonly used in diffusion model training, we adopt a modified schedule based on the observation that noise levels influence task suitability: low-level noise inputs benefit classification, while high-level noise inputs are more suited for generation. We used an inverse-cosine noise schedule~\citep{hudson2024soda} to create more appropriate training samples for both objectives.

\paragraph{Parameter-efficient training.} 
To maintain efficiency, we freeze all parameters of the pre-trained Stable Diffusion model and introduce trainable LoRA matrices within its cross-attention layers. These low-rank adaptation matrices enable fine-tuning while preserving the original model’s weights, significantly reducing the number of trainable parameters and computational overhead. We employ default LoRA settings~\citep{hu2021lora} for rank and learning rate to achieve an optimal balance between efficiency and performance without compromising generative capabilities.
\section{Experiments}
\label{sec:experiments}

\subsection{Experimental Settings}

\paragraph{Evaluation.}
We evaluate \textsc{D$^3$CL} on both image understanding and generation tasks. For understanding tasks, we use extracted features for linear probing on ImageNet-1K classification~\citep{russakovsky2015imagenet} and report top-1 accuracy. We also examine cross-dataset generalization on CIFAR-100~\citep{krizhevsky2009cifar} via few-shot transfer learning, and additionally report \emph{zero-shot} kNN classification results. Finally, to assess spatial understanding, we include a visual correspondence evaluation on SPair-71k~\citep{min2019spair}, following \citet{tang2023dift}. For generation tasks, we assess unconditional and class-conditional image generation performance on ImageNet-256 and free-form text-to-image generation on MSCOCO~\citep{lin2014mscoco}.

\paragraph{Training details.} 
We adopt pre-trained Stable Diffusion v1.4 as the base model with LoRA matrices attached to its cross-attention layers. We chose Stable Diffusion version 1.4 instead of stronger versions for fair comparison with other baselines, demonstrating that our method does not rely solely on heavily pretrained models. We trained D$^3$CL on ImageNet-1K dataset. We used features from the bottleneck layer of UNet in Stable Diffusion, processed through cross-attention for downstream classification tasks. We directly used the diffusion model output for the image generation task. Our experiments were conducted on 4 NVIDIA H100 GPUs. We trained D$^3$CL for 100 epochs using a batch size of 512 with standard image augmentation techniques.

\subsection{Evaluation Results}

\subsubsection{Image Classification}

\begin{table}[!h]
    \centering\small
    \begin{tabular}{l l l l l}
    \toprule
    \multirow{2}{*}{\textbf{Method}} & \multirow{2}{*}{\textbf{Backbone}} & \multicolumn{2}{c}{\textbf{\#Params.}} & \multirow{2}{*}{\textbf{Acc.}$\uparrow$} \\
    \cmidrule(ll){3-4}
    & & \textbf{Trainable} & \textbf{Frozen} & \\
    \midrule
    \rowcolor{lightgray} \multicolumn{5}{l}{\it contrastive based methods} \\
    SimCLR~\citep{chen2020simclr} & ResNet50$\times$2 & 94M & - & 74.1 \\
    DINO~\citep{caron2021dino} & ViT-B/16 & 86M & - & 78.0 \\
    iBOT~\citep{zhou2021ibot} & ViT-B/16 & 86M & - & 75.8 \\
    \midrule
    \rowcolor{lightgray} \multicolumn{5}{l}{\it generative based methods} \\
    MAE~\citep{he2022mae} & ViT-L/16 & 304M & - & 73.5 \\
    MAGE~\citep{li2023mage} & ViT-L/16 & 304M & 24M & \underline{78.9} \\
    GIVT$^{\dagger}$~\citep{tschannen2025givt} & ViT-L/16 & 304M & - & 65.1 \\
    \midrule
    \rowcolor{lightgray} \multicolumn{5}{l}{\it diffusion based methods} \\
    DifFeed~\citep{mukhopadhyay2023diffeed} & UNet* & 31M & 554M & 76.8 \\
    SD Features & UNet* & - & 980M & 71.8 \\
    \rowcolor{white} 
    {\bf D$^3$CL} (ours) & UNet* & 68M & 980M & \textbf{80.1} \\
    \bottomrule
    \end{tabular}
    \caption{{\bf Linear probing performance on ImageNet-1K.} We group all evaluated methods into 3 categories: contrastive based methods, generative based methods, and diffusion based methods. We directly extract features from pre-trained Stable Diffusion v1.4 model and evaluate the raw features' performance as a baseline (shown as SD Features in table). In D$^3$CL, trainable parameters refer to LoRA matrices and feature extraction module.$\dagger$ means results are from original works; * means UNet architecture from pre-trained diffusion models; SD (Stable Diffusion)}
    \label{tab:classification_main}
    \vspace{-1em}
\end{table}

\paragraph{Setup.} 
For linear probing, we attach a linear classifier to the features extracted from our frozen pre-trained models. The classifier is trained using SGD with a momentum of $0.9$, a fixed learning rate of $0.01$, and an $L_2$ regularization penalty. The linear classifier is trained on ImageNet for 50 epochs with a batch size of 256, using 20 denoising steps.

\paragraph{Results.}
Classification performance is evaluated with top-1 accuracy on the ImageNet validation set. As summarized in \Cref{tab:classification_main},  D$^3$CL outperforms the diffusion-based DifFeed~\citep{mukhopadhyay2023diffeed} by $3.3\%$. 
Furthermore, D$^3$CL surpasses contrastive-based and other generative-based methods while maintaining a significantly lower number of trainable parameters. For example, while DINO~\citep{caron2021dino} and MAE~\citep{he2022mae} require 86M / 304M trainable parameters from their ViT backbones, D$^3$CL achieves superior classification performance with 68M trainable parameters.

\begin{table}[t] 
    \begin{minipage}{0.49\textwidth}
        \centering
        \small
        \setlength{\tabcolsep}{4pt} 
        \begin{tabular}{lccc}
            \toprule
            \textbf{Method} & \textbf{Res.} & \textbf{FID} $\downarrow$ & \textbf{IS} $\uparrow$ \\
            \midrule 
            ICGAN$^{\dagger}$ & 256 & 15.6 & 59.0 \\
            ADM$^{\dagger}$ & 256 & 26.21 & 39.70 \\
            GIVT$^{\dagger}$ & 256 & 11.02 & - \\
            MAGE (ViT-L) & 256 & \underline{7.04} & \underline{123.5} \\
            \rowcolor{white} 
            {\bf D$^3$CL} (ours) & 256 & \textbf{5.56} & \textbf{142.3} \\
            \bottomrule
        \end{tabular}
        \caption{{\bf Unconditional ImageNet-256 generation.} FID computed against validation set at $256\times256$. $\dagger$ denotes results from original works.}
        \label{tab:gen_256}
    \end{minipage}
    \hfill 
    \begin{minipage}{0.49\textwidth}
        \centering
        \small
        \setlength{\tabcolsep}{4pt}
        \begin{tabular}{lccc}
            \toprule
            \textbf{Method} & \textbf{Type} & \textbf{FID} $\downarrow$ & \textbf{IS} $\uparrow$ \\
            \midrule 
            MaskGIT & MIM & \underline{6.18} & 182.1 \\
            MAGE(ViT-B) & MIM & 6.93 & \textbf{195.8} \\
            ADM & Diff. & 10.94 & 101.0 \\
            LDM & Diff. & 10.56 & 103.5 \\
            \rowcolor{white} 
            {\bf D$^3$CL} (ours) & Diff. & \textbf{5.16} & \underline{189.7} \\
            \bottomrule
        \end{tabular}
        \caption{{\bf Class-conditional ImageNet generation.} Best performance is bolded; second best is underlined.}
        \label{tab:con_gen_256}
    \end{minipage}
    \vspace{-2em}
\end{table}

\subsubsection{Visual Correspondence}

\begin{wraptable}{r}{0.38\textwidth}
    \centering
    \small
    \vspace{-12pt} 
    \begin{tabular}{lc}
        \toprule
        {\bf Method} & {\bf PCK}@bbox $\uparrow$\\
        \midrule
        DINO                     & 33.9 \\
        OpenCLIP                 & 38.4 \\
        DIFT$_{\text{sd}}$       & \underline{52.9} \\
        {\bf D$^3$CL} (ours)  & {\bf 53.0} \\
        \bottomrule 
    \end{tabular}
    \caption{{\bf PCK on SPair-71k.} }
    \label{tab:re_spair}
\end{wraptable}

Visual correspondence is a critical image understanding task used for 3D reconstruction, tracking, and segmentation. In Table~\ref{tab:re_spair}, we evaluate features extracted from D$^3$CL on semantic correspondence task to demonstrate its potential in more complex vision understanding tasks. In particular, features extracted from D$^3$CL yield better keypoint matching on SPair-71k~\citep{min2019spair} than the base pretrained diffusion model and other representation learning baselines, indicating stronger spatially grounded semantics.

\subsubsection{Image Generation}

\paragraph{Setup.} 
We evaluate our model's generative capacity through the challenging tasks of unconditional / class-conditional image generation on ImageNet. After pretraining, no additional fine-tuning is applied for image generation. The quality of the generated images is evaluated using Inception Score (IS) and Fréchet Inception Distance (FID). We generate 50k images at 256$\times$256 resolution , using 100 denoising steps per image, and calculate the metrics on the ImageNet-256 validation set.

\paragraph{Unconditional image generation.}
D$^3$CL achieves an FID of 5.56 and an IS of 142.3 on unconditional ImageNet-256 generation, indicating strong image quality and diversity. Comparative results with other state-of-the-art models are provided in \Cref{tab:gen_256}. 
These results demonstrate D$^3$CL's ability to generate diverse, high-quality images without relying on additional labeled data. This success indicates the potential of large pre-trained diffusion models in applications requiring detailed and varied image synthesis, especially in scenarios where explicit class labels are unavailable.

\paragraph{Class-conditional image generation.} 
For direct class-label conditioned generation, we adopt a conditional encoder similar to \citet{rombach2022ldm} consisting of a single learnable embedding layer with a dimensionality of 512. We assess D$^3$CL's conditional generation performance on the ImageNet-1K validation set and compare it against baseline methods, with results summarized in \Cref{tab:con_gen_256}.
D$^3$CL achieves a significantly improved FID score of 5.16, indicating superior image quality and diversity compared to baseline models. Furthermore, it attains a high IS of 189.7, closely matching the top-performing MAGE model (195.8), demonstrating its effectiveness in class-conditional generation. The slight difference in the IS score may stem from D$^3$CL's pretraining on large-scale datasets with distributions differing from ImageNet, which is used for IS evaluation. 
Overall, these results underline D$^3$CL’s effectiveness in balancing image fidelity and semantic alignment.

\begin{figure}[h!]
    \centering
    \small
    \begin{minipage}[t]{0.5\textwidth}
        \vspace{4pt} 
        \centering
        \setlength{\tabcolsep}{3pt} 
        \begin{tabular}{lccc}
            \toprule
            {\bf Method} & {\bf Trained MSCOCO} & {\bf FID$\downarrow$} & {\bf CLIP$\uparrow$} \\
            \midrule
            U-Net~\citep{ronneberger2015unet}    & \checkmark & 18.73       & 79.41 \\ [.3em]
            LDM~\citep{rombach2022ldm}           & \ding{55}  & 23.31       & 84.65 \\ [.3em]
            SD v1.4                              & \ding{55}  & 20.52       & 88.10 \\ [.3em]
            {\bf D$^3$CL} (ours)              & \ding{55}  & {\bf 16.37} & {\bf 92.45} \\
            \bottomrule
        \end{tabular}
        \vspace{1.5em}
        \captionof{table}{{\bf MSCOCO-256 Text-to-Image.} FID computed from 40K samples. \ding{55} indicates pre-trained on larger datasets (LAION/ImageNet) but not MSCOCO.}
        \label{tab:re_t2i}
    \end{minipage}
    \hfill 
    \begin{minipage}[t]{0.4\textwidth}
        \vspace{0pt} 
        \centering
        \begin{tikzpicture}
            \begin{axis}[
                width=\linewidth, 
                height=0.7\linewidth, 
                xlabel={Training Iteration},
                ylabel={FID-50K},
                xlabel style={font=\bfseries\small},
                ylabel style={font=\bfseries\small},
                tick label style={font=\footnotesize},
                xmode=log,
                ymode=log,
                log basis x=10,
                log basis y=10,
                xmin=1e4, xmax=4e5,
                ymin=5, ymax=120,
                xtick={1e4,5e4,1e5,4e5},
                xticklabels={10K,50K,100K,400K},
                ytick={112.3,70,35,15,10,6.29},
                yticklabels={112,70,35,15,10,6.3},
                grid=both,
                grid style={dashed, gray!60},
                axis line style={black, thick},
                legend style={
                    at={(0.98,0.98)},
                    anchor=north east,
                    draw=gray!60,
                    fill=white,
                    font=\scriptsize,
                },
            ]
            \addplot[very thick, mark=*, mark size=2pt, color=cyan] coordinates {
                (1e4, 112.3) (5e4, 52.3) (1e5, 19.4) (4e5, 7.9)
            }; \addlegendentry{REPA}
            
            \addplot[very thick, mark=*, mark size=2pt, color=red] coordinates {
                (1e4, 90.7) (5e4, 42.2) (1e5, 14.9) (4e5, 6.1)
            }; \addlegendentry{D$^3$CL}
            \end{axis}
        \end{tikzpicture}
        \vspace{-0.2em}
        \captionof{figure}{{\bf Training efficiency.} D$^3$CL converges faster than REPA~\cite{yu2025repa} on ImageNet.}
        \label{fig:train_efficiency}
    \end{minipage}
    \vspace{-1.5em}
\end{figure}

{\bf Text-to-image generation.}
We evaluate text-to-image generation using MSCOCO captions and standard CLIP-based metrics alongside FID, as shown in Table~\ref{tab:re_t2i}. We apply text encoder from SD v1.4 for text context and add image context from image encoder output from Gaussian noise input. These results verify that our joint objectives and LoRA adaptation do not degrade free-form prompt image generation: D$^3$CL achieves 92.45 CLIP score, compared to 88.10 for the SD v1.4 baseline, indicating improved prompt adherence.

\subsubsection{Transfer Learning on Classification}

\begin{table}[!h]
\centering\small
\begin{tabular}{l l l l l l l }
\toprule
\textbf{Method} & \textbf{Type} & \textbf{Backbone} & \textbf{\#Params.} & \textbf{Acc.@25}$\uparrow$ & \textbf{Acc.@0}$\uparrow$\\
\midrule
SimCLR~\citep{chen2020simclr} & Contrastive & ResNet50$\times$2 & 94M/\color{gray}{-} & 58.9 & 52.3 \\
MAGE~\citep{li2023mage} & Generative & ViT-L/16 & 304M/\color{gray}{24M} & \underline{72.0} & \underline{63.5}\\
DifFeed~\citep{mukhopadhyay2023diffeed} & Diffusion & UNet* & 31M/\color{gray}{554M} & 70.3 & 61.8\\
\rowcolor{white} 
\textbf{D$^3$CL}(ours) & Diffusion & UNet* & 68M/\color{gray}{980M} & \textbf{73.1} & {\bf 65.2}\\
\bottomrule
\end{tabular}

\caption{{\bf Transfer learning performance on CIFAR-100.} Top-1 accuracy of transfer learning on CIFAR-100 dataset of models pretrained on ImageNet-1K is reported. We choose one baseline method from each of our three groups of methods listed in \Cref{tab:classification_main} in the main paper. D$^3$CL maintains the best performance over three baselines. We present the number of both trainable\textcolor{gray}{/frozen} parameters in \textbf{``\#Params.''} column. * means UNet architecture from pre-trained diffusion models.}
\label{tab:ablation_classification}
\vspace{-1em}
\end{table}

\paragraph{Few-shot learning.} To evaluate the generalization ability of D$^3$CL, we measure its performance on the CIFAR-100 dataset under a low-data regime, where only 25 samples per class are used for training. As shown in \Cref{tab:ablation_classification}, D$^3$CL outperforms all selected baseline methods, demonstrating its robustness in low-data regimes. These results highlight its capacity to extract meaningful representations and maintain strong performance even with limited training data.

{\bf Zero-shot learning.}
To further isolate representation quality from supervised adaptation, we additionally evaluate \textbf{zero-shot} performance on CIFAR-100 with kNN classification. As shown in Table~\ref{tab:ablation_classification}, D$^3$CL outperforms baselines on zero-shot kNN accuracy, consistent with the few-shot results, indicating that the learned diffusion representations generalize beyond ImageNet fine-tuning and remain effective on a different domain.

\subsection{Ablation Study}

\paragraph{Ablation of individual components in D$^3$CL.}
\Cref{tab:ablation_structure} illustrates the contribution of each component to the performance of D$^3$CL, starting from the pre-trained Stable Diffusion v1.4 baseline, which achieves 71.8\% accuracy using the direct bottleneck layer output for linear probing.
i) Adding an attention-based feature extraction network improves accuracy to 74.3\% (+2.5\%).
ii) Incorporating LoRA training further boosts accuracy to 78.0\% (+3.7\%), with only the reconstruction objective applied. 
iii) Finally, adding a contrastive loss achieves an accuracy of 80.1\% (+2.1\%). 
Overall, D$^3$CL demonstrates an 8.3\% improvement over baseline (SD v1.4), with LoRA and contrastive loss providing a significant boost for optimal performance.
We additionally report a concise breakdown for inference latency caused by each component.

\paragraph{Impact of contrastive loss via weighting parameter $\lambda$.}
Our ablation study examines the influence of the contrastive loss during training, as shown in Table~\ref{tab:re_ablation}.
Experiment shows that $\lambda=0.1$ provides the best overall performances on both tasks. 
We noticed that increasing $\lambda$ does not always lead to improved linear probing accuracy, which supports our unified framework: the combined loss benefits both tasks. 
The reconstruction loss acts as a regularizer for the classification task, meaning that increasing $\lambda$ may not necessarily improve linear probing performance. Therefore, our default $\lambda$ is chosen to balance performance across both tasks.

\begin{table*}[h]
    \centering
    \small
    \begin{minipage}[t]{0.6\linewidth}
        \centering
        \vspace{0pt} 
        \setlength{\tabcolsep}{3pt}
\begin{tabular}{lll}
    \toprule
    \textbf{Component}   & \textbf{Inference Latency} & \textbf{Acc.$\uparrow$}  \\
    \midrule
    SD v1.4              & 1.858$\pm$0.008                     & 71.8                 \\
    \arrayrulecolor{black!40}\midrule
    + Feature Extraction & 1.861$\pm$0.011 ($+0.1\%$)          & 74.3 (+2.5)               \\
    + LoRA Training      & 2.094$\pm$0.019 ($+12.0\%$)         & 78.0 (+3.7)               \\
    + $\mathcal{L}_\text{contrast}$ (\textsc{D$^3$CL})  & 2.094$\pm$0.019 ($+12.0\%$) &  80.1 (+2.1)\\
    \bottomrule
    \end{tabular}
        \captionof{table}{{\bf Ablation of \textsc{D$^3$CL} components on linear probing accuracy and inference latency.} Contrastive loss significantly improves accuracy. 
        Inference latency is based on 100 steps generation.}
        \label{tab:ablation_structure}
    \end{minipage}
    \hfill
    \begin{minipage}[t]{0.35\linewidth}
        \centering
        \vspace{0pt} 
        
        \begin{subtable}[t]{\linewidth}
        \centering
        \begin{tabular}{p{0.40\linewidth} p{0.20\linewidth} p{0.20\linewidth}}
            \toprule
            {\bf $\lambda$}  & {\bf FID$\downarrow$} & {\bf Acc.$\uparrow$} \\ [.25em]
            \midrule
            0                & 14.71      & 78.0       \\
            $1e^{-3}$        & 12.32      & 79.7       \\
            $1e^{-1}$        & {\bf 5.56} & {\bf 80.1} \\
            1                & 5.45       & 78.3       \\
            \bottomrule
        \end{tabular}
    \end{subtable}
        
        \vspace{5pt}
        \captionof{table}{{\bf Ablation study on Loss weight $\lambda$.} We evaluate top-1 kNN accuracy and FID on ImageNet-1K. Default parameters are bolded.}
        \label{tab:re_ablation}
    \end{minipage}
\end{table*}
\vspace{-1.5em}

\subsection{Discussion}

\paragraph{Efficiency analysis.}
To improve computational efficiency, our model minimizes the number of trainable parameters while maintaining competitive performance. 
\Cref{tab:classification_main} compares the total number of trainable parameters across different models.
As demonstrated, D$^3$CL reduces the number of trainable parameters by 28\% compared to SimCLR and 78\% compared to MAGE.
In Fig.~\ref{fig:train_efficiency}, we report FID convergence during training iteration compared with REPA when trained from scratch on a SiT model~\cite{ma2024sit}. D$^3$CL consistently achieves lower FID-50K than REPA at during 10K–400K iterations, with a clear gap already visible in early training.

\begin{figure*}[h!]
    \centering
    \includegraphics[trim={50 0 50 0},clip,width=0.95\linewidth]{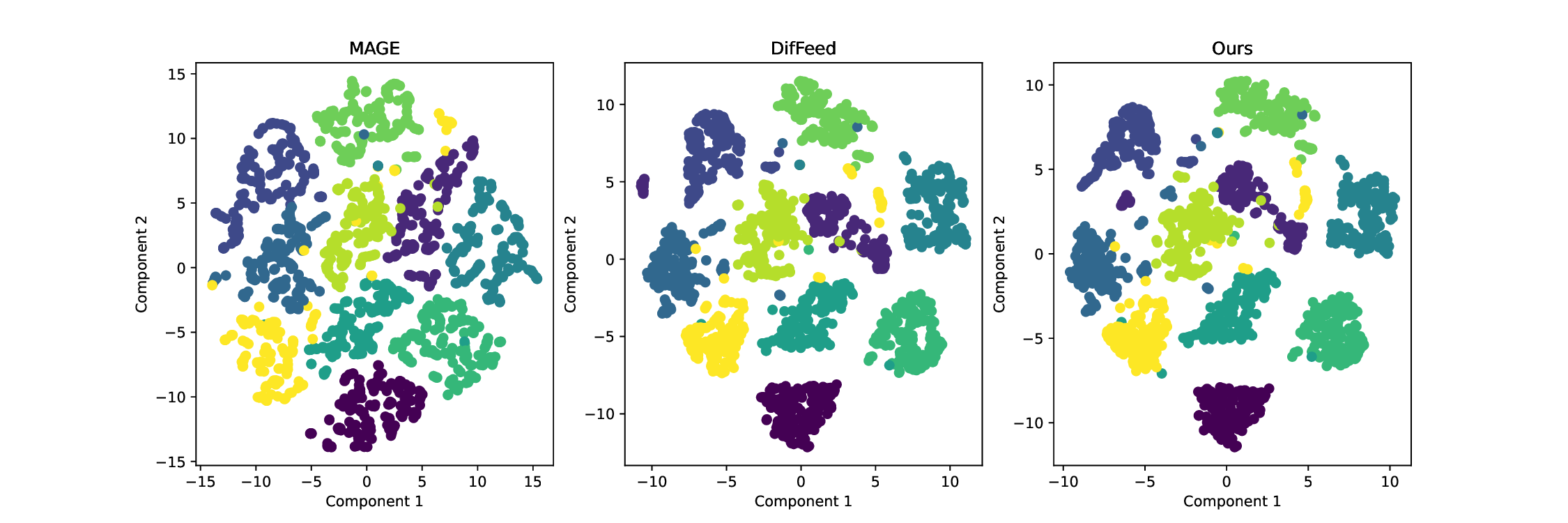}
    \caption{{\bf D$^3$CL produces more separable feature embeddings.} By t-SNE visualization on feature linear separability, we compare D$^3$CL \emph{(unified method)} against MAGE~\citep{li2023mage} \emph{(unified method)}, which also uses a contrastive-based loss, and DifFeed~\citep{mukhopadhyay2023diffeed} \emph{(generative method)}, which elucidates feature extraction method design on a frozen diffusion model.}
    \label{fig:tsne}
\end{figure*}
\vspace{-0.5em}

\paragraph{D$^3$CL's feature representations.} 
We use t-SNE to visualize feature vectors from the ImageNet validation set extracted by MAGE, DifFeed, and our method, as shown in \Cref{fig:tsne}. We observed that our method produces more distinct and well-separated clusters, affirming the discriminative strength of our model's features.
\section{Conclusion}
\label{sec:conclusion}

In this paper, we introduced D$^3$CL, a novel framework that efficiently adapts pretrained diffusion models for both generative and discriminative tasks within a unified framework. 
By combining reconstruction and contrastive losses and utilizing varying noise levels to balance the demands of both tasks, D$^3$CL demonstrates strong performance and enhanced computational efficiency.
%
Our extensive evaluation highlights the framework's potential to address critical challenges in the field of self-supervised learning and generative modeling, such as fast adaptation of pretrained diffusion models to a variety of discriminative tasks. A promising direction for future research would be the extension of D$^3$CL to integrate additional tasks and modalities.

\newpage



\bibliography{ref}
\bibliographystyle{plainnat}

\newpage
\appendix

\setcounter{page}{1}

\section*{Supplementary Material Overview}
\label{sec:supp_overview}

This supplementary material provides additional insights into our method, including detailed implementation specifications and visualization. For implementation, we visualize the inference pipeline of \textsc{D$^3$CL}. Visualization presents additional visualizations on generation results.

\section{Experiment Details}

\begin{table}[h]
    \centering
    \begin{minipage}{0.49\textwidth}
        \centering
        \begin{tabular}{l|l}
        \toprule
        \textbf{Hyperparameter} & \textbf{Value} \\
        \midrule optimizer & AdamW \\
        learning rate & $1.5 \times 10^{-4}$ \\
        momentum & $0.9$ \\
        weight decay & $0.05$ \\
        batch size & $512$ \\
        learning rate schedule & cosine decay\\
        training epochs & $100$ \\
        warmup epochs & $5$ \\
        \bottomrule
        \end{tabular}
        \caption{\textbf{Pretraining settings.}}
        \label{tab:param_training}
    \end{minipage}
    \hfill
    \begin{minipage}{0.49\textwidth}
        \centering
        \begin{tabular}{l|l}
        \toprule
        \textbf{Hyperparameter} & \textbf{Value} \\
        \midrule optimizer & SGD \\
        learning rate & $0.01$ \\
        momentum & $0.9$ \\
        weight decay & $0.05$ \\
        batch size & $256$ \\
        learning rate schedule & cosine decay\\
        training epochs & $50$ \\
        warmup epochs & $5$ \\
        \bottomrule
        \end{tabular}
        \caption{\textbf{Linear probing settings.}}
        \label{tab:param_lp}
    \end{minipage}
\end{table}

\paragraph{Hyperparameter settings.} \Cref{tab:param_training,tab:param_lp} summarize the hyperparameter settings used in the D$^3$CL framework. 
Pretraining uses AdamW optimizer~\citep{loshchilov2017adamw}  with a base learning rate of $1.5 \times 10^{-4}$, momentum of $0.9$, and weight decay of $0.05$. The batch size is set to $512$, and training follows a cosine decay learning rate schedule over $100$ epochs, with $5$ warmup epochs. 
Linear probing applies SGD optimizer with a learning rate of $0.01$, momentum of $0.9$, and weight decay of $0.05$. The batch size is $256$, using a cosine decay learning rate schedule for $50$ training epochs, including $5$ warmup epochs. 
Image generation employs classifier-free guidance (CFG) with $100$ diffusion steps as the default setting.

\paragraph{Latent encoder and decoder.} 
These components encode input images into a compact latent space and decode them back into images. Leveraging this compressed latent space reduces computational overhead while facilitating efficient feature extraction. We applied the pre-trained VAE used in Stable Diffusion~\citep{rombach2022ldm} with a down-sampling factor $f=4$ as our default encoder/decoder.

\paragraph{Conditional encoder.} 
D$^3$CL incorporates a conditional mechanism based on image inputs, encoding them into conditioning tokens. Similar to \citet{rombach2022ldm}, we used a transformer-based conditional encoder with an embedding dimension of 512. However, instead of directly applying the pre-trained conditional encoder from \citet{rombach2022ldm}, we adapted it to address the modality difference between text prompts and image inputs. The resulting embeddings are integrated into the denoising UNet through cross-attention layers, facilitating effective conditioning during the generation process.

\paragraph{LoRA weight matrices.} LoRA~\citep{hu2021lora} matrices efficiently adapt large pre-trained models by introducing trainable low-rank matrices to specific layers. In our implementation, we apply LoRA matrices to the cross-attention layers, allowing the model to tailor its responses to inputs from different modalities with minimal added parameters. This approach preserves the core features learned by the pre-trained Stable Diffusion model while optimizing performance for new tasks. For our setup, we applied LoRA matrices with a rank of 16.

\section{Inference Pipeline}

\begin{figure*}[htbp]
    \centering
    \includegraphics[width=\linewidth]{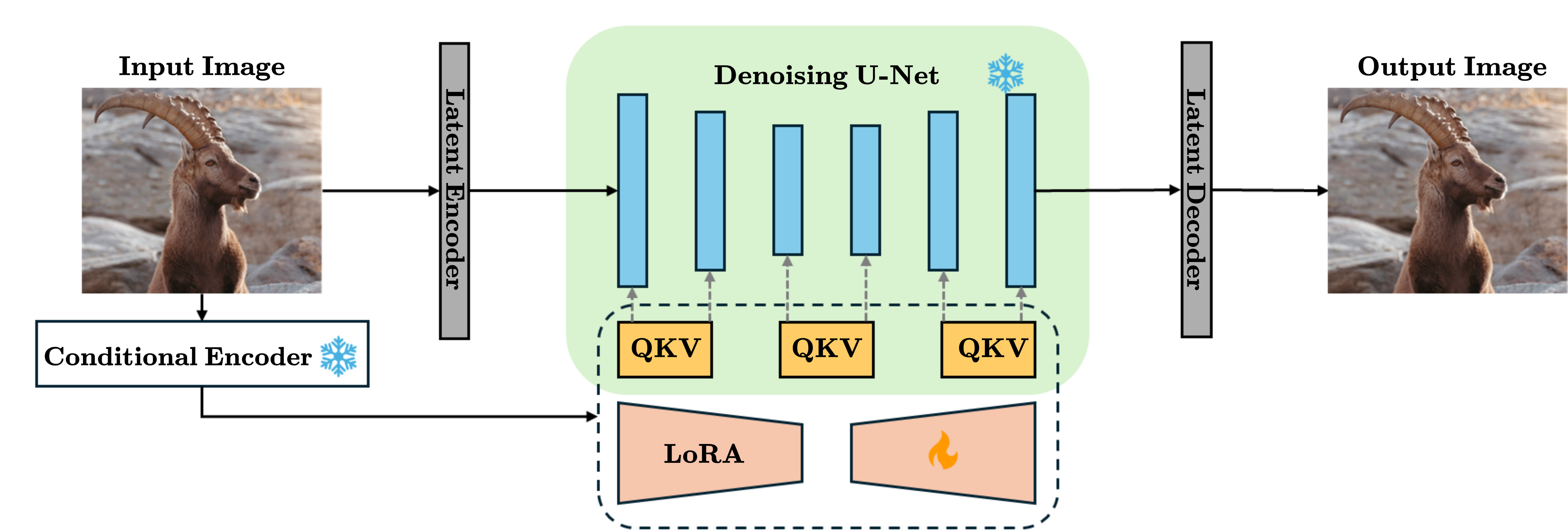}
    \caption{\textbf{Unified inference for classification and generation.} The condition encoder processes an input image to produce a feature representation, which serves as conditional latent for the denoising UNet. In the classification task, the feature map output from the attention head is used to predict class labels. For the generation task, this conditional latent guides the synthesis of coherent images according to the given input conditions. For unconditional image generation, a Gaussian noise image is used as conditional input.}
    \label{fig:method_inference}
\end{figure*}

As shown in \Cref{fig:method_inference}, for the classification task, the feature map from the attention head predicts class labels. For generation, this feature guides image synthesis based on conditions. For unconditional generation, a pure Gaussian noise image ( $T=T_{max}$) is used as input.

\section{Visualization}
\label{sec:vis}

\begin{figure}[htbp]
    \centering
    \includegraphics[width=0.6\linewidth]{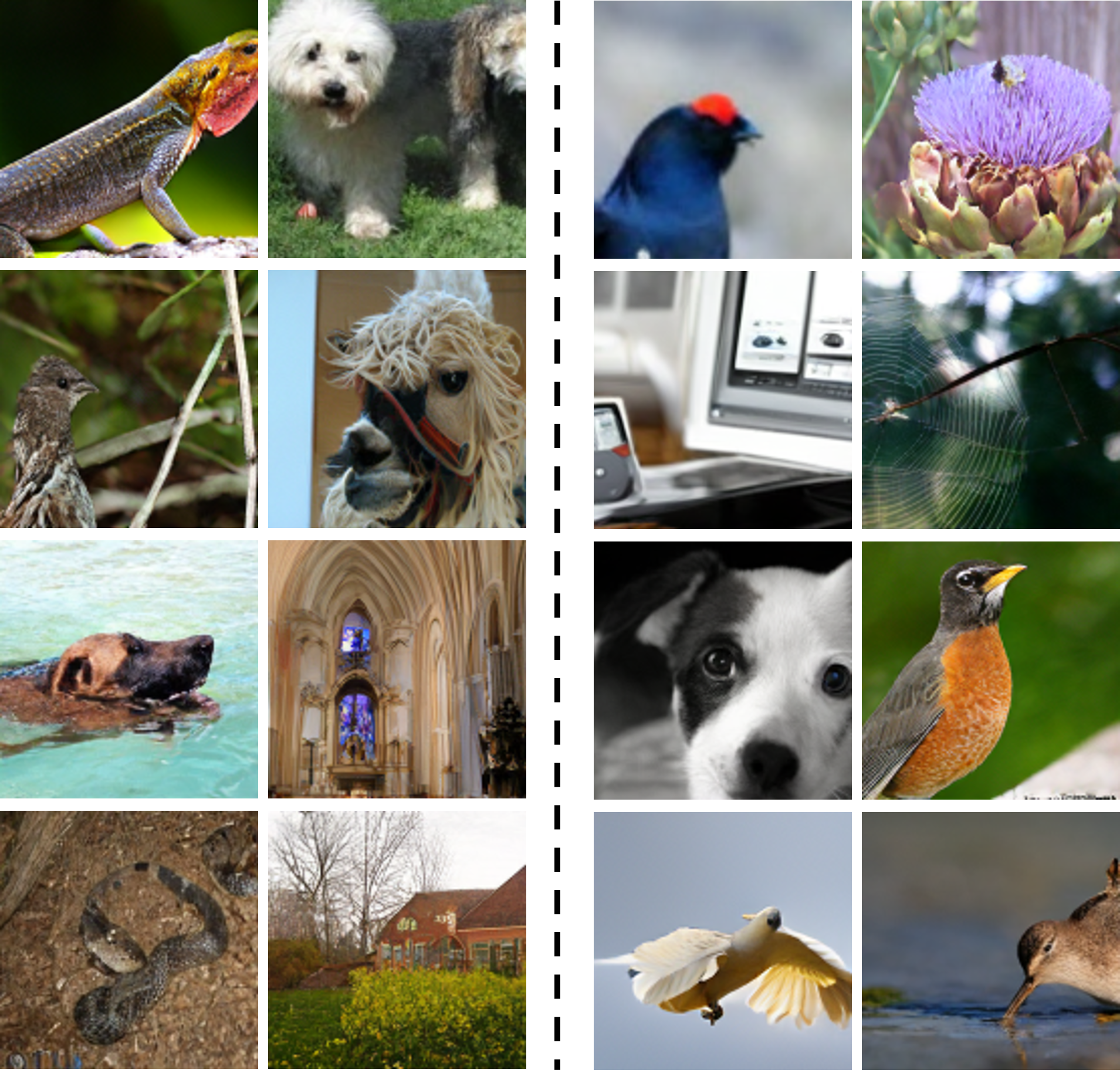}
    \caption{{\bf D$^3$CL improves ImageNet sample quality.} Generated images from MAGE~\citep{li2023mage} pre-trained ViT-B/16 model (left) and \textsc{D$^3$CL} (right). We employ unconditional generation on ImageNet. D$^3$CL brings images with more vivid details, illustrating its strong performance on generating high-fidelity images.}
    \label{fig:img_gen}
\end{figure}

We show some of the generated results by D$^3$CL and compare them with outputs from pre-trained MAGE ViT-B/16 model, as illustrated in \Cref{fig:img_gen}.

\end{document}